\documentclass[runningheads]{llncs}
\usepackage[T1]{fontenc}
\usepackage[table,xcdraw]{xcolor}
\usepackage{hyperref}
\usepackage{booktabs}
\usepackage{multirow}
\usepackage{multicol}
\usepackage{caption}
\usepackage{graphicx}
\usepackage{subcaption}
 \usepackage{amsmath} 
\usepackage{academicons}
\usepackage{svg}

\newcommand{\orcid}[1]{\href{https://orcid.org/#1}{\includegraphics[width=0.03\textwidth]{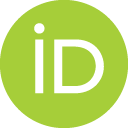}}}

\begin{document}
%
%\title{GCNNs on FPGA for DAS data processing}
\title{Hardware-Accelerated Event-Graph Neural Networks for Low-Latency Time-Series Classification on SoC FPGA}

\titlerunning{Hardware-Accelerated Event-Graph Neural Networks}
% This is for the Springer version

%\author{Hiroshi Nakano\inst{1}\orcidID{0009-0009-5493-4689} \and
%Krzysztof Blachut\inst{2}\orcidID{0000-0002-1071-335X} \and
%Kamil Jeziorek\inst{2}\orcidID{0000-0001-5446-3682} \and
%Piotr Wzorek\inst{2}\orcidID{0000-0003-3885-600X} \and
%Manon Dampfhoffer\inst{3}\orcidID{0000-0002-2622-3101} \and
%Thomas Mesquida\inst{3}\orcidID{0000-0002-2572-5353} \and
%Hiroaki Nishi \inst{1}\orcidID{0000-0002-6331-2947} \and
%Tomasz Kryjak\inst{2}\orcidID{0000-0001-6798-4444} \and
%Thomas Dalgaty\inst{3}\orcidID{0000-0003-0326-2121}}

% This is for the arxiv version

% REMOVED FOR BLIND REVIEW
\author{Hiroshi Nakano\inst{1,3}\orcid{0009-0009-5493-4689} \and
Krzysztof Blachut\inst{2}\orcid{0000-0002-1071-335X} \and
Kamil Jeziorek\inst{2}\orcid{0000-0001-5446-3682} \and
Piotr Wzorek\inst{2}\orcid{0000-0003-3885-600X} \and
Manon Dampfhoffer\inst{3}\orcid{0000-0002-2622-3101} \and
Thomas Mesquida\inst{3}\orcid{0000-0002-2572-5353} \and
Hiroaki Nishi \inst{1}\orcid{0000-0002-6331-2947} \and
Tomasz Kryjak\inst{2}*\orcid{0000-0001-6798-4444} \and
Thomas Dalgaty\inst{3}\orcid{0000-0003-0326-2121}}

\authorrunning{H. Nakano et al.}
\institute{Graduate School of Science and Technology, Keio University, Tokyo, Japan \\
\email nakano@west.sd.keio.ac.jp, west@keio.jp \\
\and Embedded Vision Systems Group, Computer Vision Laboratory, \\AGH University of Krakow, Poland \\
\email \{kblachut,kjeziorek,pwzorek,tomasz.kryjak\}@agh.edu.pl \\
\and CEA-List, Université Grenoble Alpes, Grenoble, France \\
\email \{Manon.DAMPFHOFFER,thomas.mesquida,Thomas.DALGATY\}@cea.fr \\
*Corresponding author: tomasz.kryjak@agh.edu.pl}
\maketitle              % typeset the header of the contribution
\begin{abstract}
%As the quantities of data recorded by embedded edge sensors grows, so too does the need for intelligent local processing. 
%Such data often comes in the form of time-series signals, based on which real-time predictions could be made locally using an AI model. 
%However, a hardware-software approach capable of making low-latency predictions with a low-power consumption is required. 
%In this paper we present, for the first time, a hardware implementation of an event-graph neural network for time-series prediction. 
%We leverage the dynamic audio sensor to convert the input time-series signals into a sparse event-data format that allows the event-graph to drastically reduce the number of calculations relative to other AI methods.
%We implement the design on a Versal VCK190 FPGA and apply it to the real-time processing of the spiking Heidelberg digits dataset in order to benchmarking our approach against FPGA-based spiking neural networks.
%We find that, due to a number of novel optimisations, we are able to achieve floating-point equivalent accuracy and significantly outperform the spiking neural network implementations using less resources and with a reduced power-consumption.

As the quantities of data recorded by embedded edge sensors grow, so too does the need for intelligent local processing. Such data often comes in the form of time-series signals, based on which real-time predictions can be made locally using an AI model. However, a~hardware-software approach capable of making low-latency predictions with low power consumption is required. In this paper, we present a~hardware implementation of an event-graph neural network for time-series classification. We leverage an artificial cochlea model to convert the input time-series signals into a~sparse event-data format that allows the event-graph to drastically reduce the number of calculations relative to other AI methods. We implemented the design on a~SoC FPGA and applied it to the real-time processing of the Spiking Heidelberg Digits (SHD) dataset to benchmark our approach against competitive solutions.
Our method achieves a~floating-point accuracy of 92.7\% on the SHD dataset for the base model, which is only 2.4\% and 2\% less than the state-of-the-art models with over 10$\times$ and 67$\times$ fewer model parameters, respectively. It also outperforms FPGA-based spiking neural network implementations by 19.3\% and 4.5\%, achieving 92.3\% accuracy for the quantised model while using fewer computational resources and reducing latency.

%In this work we implement GCNNs for DAS data processing for FPGA. 
%We take advantage of hardware-aware design method and perform a number of ablation studies to achieve most efficient audio processing system (why?). 

\end{abstract}
%
%
%
% -------------------------------------------------------------------------------------------------------------------
\section{Introduction}
%\todo{check "frequency","channel","frequency radius","channel radius","frequency channel" and unify}

\begin{figure}[!h]
    \centering
    \includegraphics[width=1\linewidth]{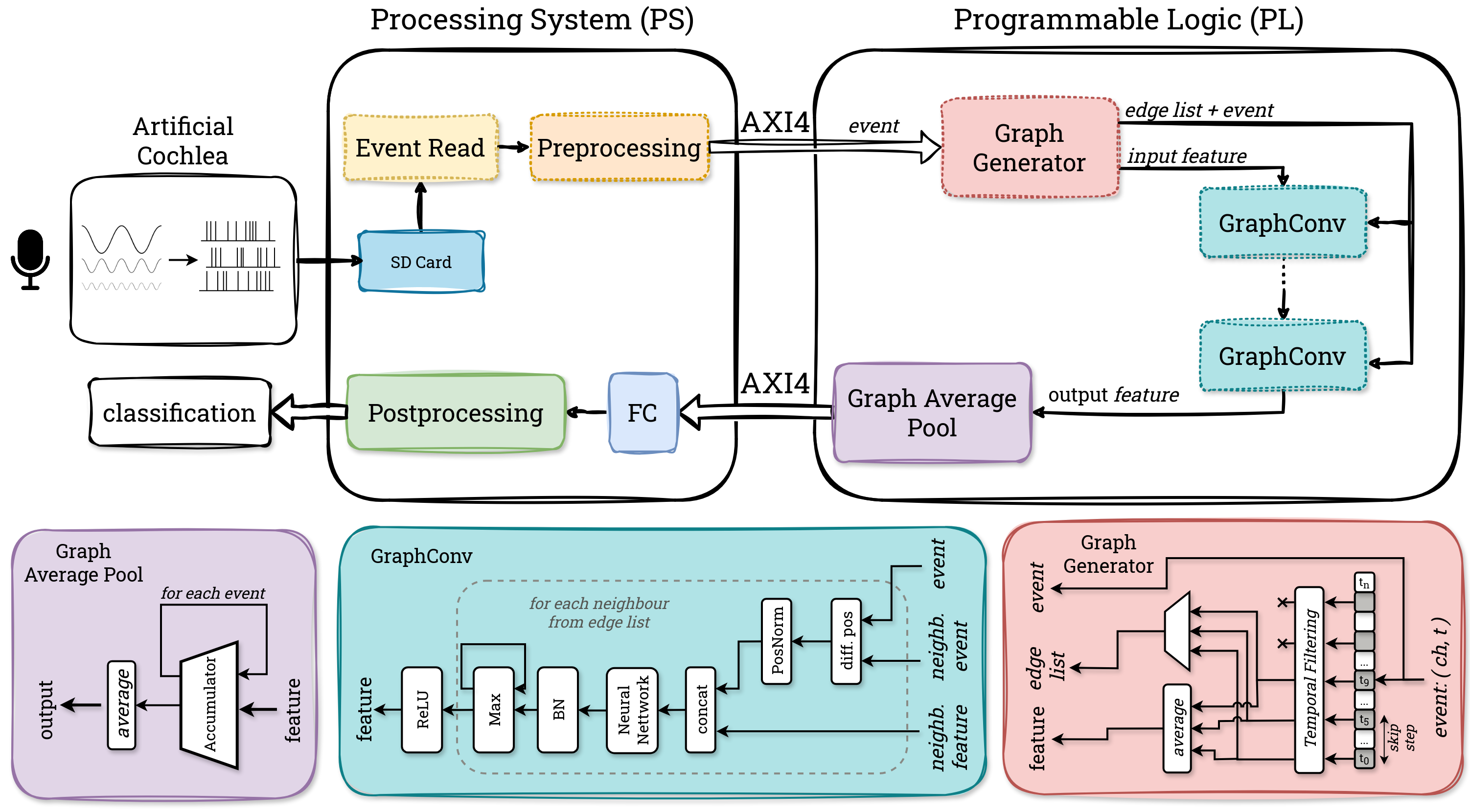}
    \caption{Overview of our hardware-accelerated event-graph neural network implementation. %Events from the Artificial Cochlea are first read and preprocessed in the PS, then processed by an asynchronous graph neural network in the PL.
    Dotted modules indicate event-by-event operation.
    %, where the Graph Average Pool updates the accumulator asynchronously and returns outputs to the fully-connected (FC) classifier in PS part after processing all events.
    For testing, we read data from an SD card and simulate real intervals between events.}
    \label{fig:teaser}
\end{figure}

As the Internet of Things expands, distributed sensors are collecting ever-increasing quantities of data. 
This has driven the need for efficient edge computing systems capable of processing this data locally, for example, to make predictions \cite{ren2023deep}. 
The energy consumption and latency of these systems are of particular importance. 

In most cases, the raw data produced by the sensors is time-series -- continuous signals following the evolution of environmental variables \cite{lim2021time}. For example, sensors that monitor the vibration of mechanical parts have been used to predict failures in gearboxes \cite{saufi2020gearbox}, or implantable cardioverter-defibrillators monitor the state of a~patient's heart in order to apply an electric shock in the event of dangerous fibrillation \cite{dimarco2003implantable}. 

It is becoming increasingly common to use AI methods to process this time-series data. However, using conventional hardware, such as GPUs (graphics processing units), consumes too much power, and microprocessors may struggle to meet the latency requirements of many applications. 
FPGAs (field programmable gate arrays) and custom integrated circuits offer a~means of implementing architectures optimised to specific AI methods that are capable of meeting latency requirements while minimising power consumption \cite{Al-Ameri,Al-Ali,Guo}.

% Furthermore, it is necessary to apply the model at periodic intervals, and important temporal information present in the signal below the sampling frequency cannot be leveraged.

A~particularly promising method is event-based AI, which operates on the sparse data generated by event-based sensors and allows reducing power consumption and prediction latency \cite{gallego2020event}. Event-based sensors, instead of regularly sampling an environmental variable, generate ``events'' in case of the changes in the signal. Event-based time-series data is generated by a~class of sensors known as artificial cochleas (AC) (also referred to as dynamic audio sensors  or silicon cochleas) \cite{liu2013asynchronous,mostafa2024}. %, which are a specific type of an Artificial Cochlea model. 
Their operating principle is to apply a~bank of band-pass filters to separate the signal into multiple frequency channels. A~digital pulse (i.e. an event) is generated per channel in an asynchronous manner when the signal intensity changes by a~pre-defined threshold. This results in a~sparse spectrogram, %(Fig. \ref{fig:visualisation})
which an event-based AI method exploits to perform efficient computation. 

Processing AC-generated event-data has been the subject of many publications. 
The most common approach is to apply spiking neural networks (SNNs) implemented for specialised hardware \cite{ortner2023online,carpegna2024spiker,dalgaty2024mosaic}. However, it is not clear how event-data sparsity can be truly exploited due to the nondeterministic pattern of synaptic weight-memory access \cite{dampfhoffer_tetci_2022,dalgaty2023cnn} inherent to SNNs. 

Recently, event-graph neural networks have been proposed as an alternative way of processing event-data \cite{li2021graph,dalgaty2023hugnet,mesquida2023g2n2,jeziorek2023memory,lars}.
The event-graph approach consists of a~dynamically updated graph generated by an event-sensor, and it involves applying graph convolutions on the resulting data structure.
Unlike SNNs, the weight access pattern for many event-graph models is deterministic. This may provide an opportunity to develop new event-based AI hardware that is truly capable of exploiting the inherent sparsity of data to reduce power consumption and latency. Although digital architectures for accelerating event-graphs have been proposed in the context of computer vision \cite{jeziorek2024embedded,yang2024evgnn}, a~dedicated architecture for time-series audio applications has not yet been considered. 

In this paper, we propose, for the first time, a~hardware accelerator implemented on a~SoC FPGA device for event-graph AC data classification (Figure \ref{fig:teaser}). %for the first time a system on-chip, fully implemented on FPGA, comprising an event-graph accelerator for prediction applications based on event-based Dynamic Audio Sensors. 
Specifically, we consider a~recently proposed spectro-temporal model \cite{lars} developed for keyword spotting and evaluated on the Spiking Heidelberg Digits (SHD) dataset \cite{cramer2020}, which is a~representative for time-series data.
% It is detailed how this software model must be adapted in order to be realised efficiently on the FPGA. We compare our implementation to existing SNN FPGA implementations targetted towards the same benchmark and additionally perform a synthesis in an advanced CMOS node in order to estimate power-consumption of a potential ASIC implementation. 

% In the introduction, we introduce Dynamic Audio Sensors (DAS), shortly discuss methods of their processing. 

% We emphasise the motivation behind the acceleration of audio processing (\textbf{Why do we need real-time, low-latency and energy-efficient embedded systems for audio processing?}).

% MOTIVATION:
% - low - latency (why we need to know the digit so fast)
% - low - power
% - 

% Event-by event operation. 
% More general, not only DAS.
% Time series - fast audio processing - like industrial process surveillance.

% We indicate, that in this paper we optimize work from "Event-based Audio Prediction with Spectro-Temporal Event-Graphs" with methods form "Embedded Graph Convolutional Networks for Real-Time Event Data Processing on SoC FPGAs" \cite{DVS-FPGA}. 

We summarise our main contribution as follows:

\begin{itemize}
\item  We use the hardware-aware design method to propose optimisations required to implement spectro-temporal event-graphs in reconfigurable hardware with low power, low latency and low resource utilisation. 

\item  We propose the first embedded system for event-graph-based audio processing on a~SoC FPGA and also the first hardware implementation that supports fully asynchronous event-by-event processing with conservation of data's temporal sparsity.

\item We achieve a~new state-of-the-art performance for FPGA solutions applied to the Spiking Heidelberg Digits dataset, demonstrating significant improvements in resource utilisation, latency reduction and accuracy compared to previous SNN-based approaches.
\end{itemize}

The remainder of this paper is organised as follows.
In Section \ref{sec:related_work} we present an overview of the related work.
In Section \ref{sec:proposed_system} we introduce the proposed embedded audio processing system.
In Section \ref{sec:evaluation} we provide hardware implementation details and present the results of ablation studies.
We conclude with a~discussion on future research directions.

% --------------------------------------------------------------------------
\section{Related work}
\label{sec:related_work}

%\textbf{We cite some works on "event-like" data processing (may be DVS), with strong emphasis on GCNNs.}

%\textbf{We introduce some works (outline SOTA) on event-based audio processing (focusing on "Event-based Audio Prediction with Spectro-Temporal Event-Graphs")}

%\textbf{We discuss FPGA platform's advantages and limitations.}
%\textbf{We introduce FPGA-implemented systems for events and point-clouds (focusing on "Embedded Graph Convolutional Networks for Real-Time Event Data Processing on SoC FPGAs").}

%\textbf{Here we also need to introduce our dataset and describe it. }
\subsection{Event-Audio Data}

\begin{figure}[htb]
    \centering
    \begin{minipage}{0.47\textwidth}
        \centering
        \includegraphics[width=\textwidth]{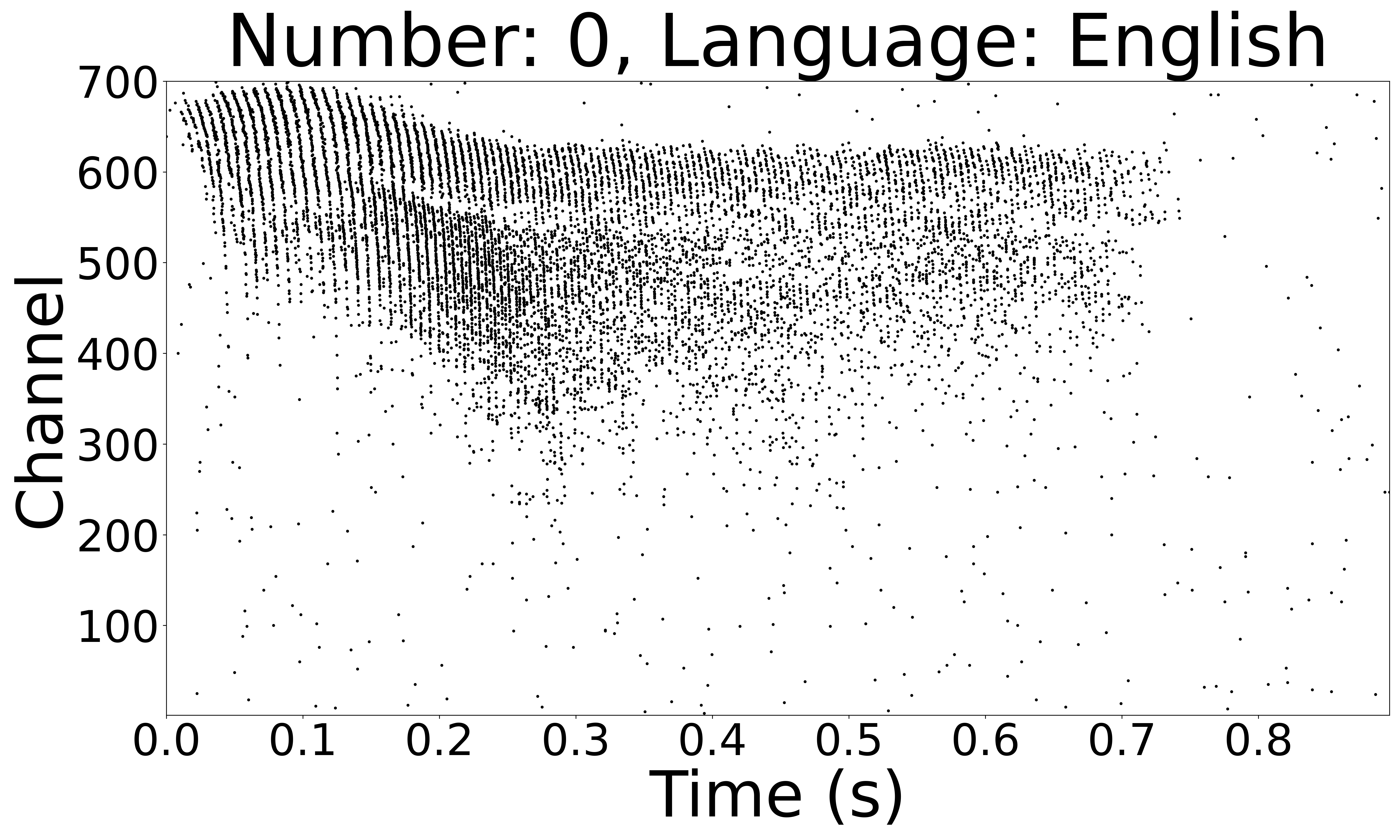}
    \end{minipage}
    % \hfill
    % \hspace{1cm}
    % \begin{minipage}{0.32\textwidth}
    %     \centering
    %     \includegraphics[width=\textwidth]{shd_spike_raster_German_6.png}
    % \end{minipage}
    \begin{minipage}{0.47\textwidth}
        \centering
        \includegraphics[width=\textwidth]{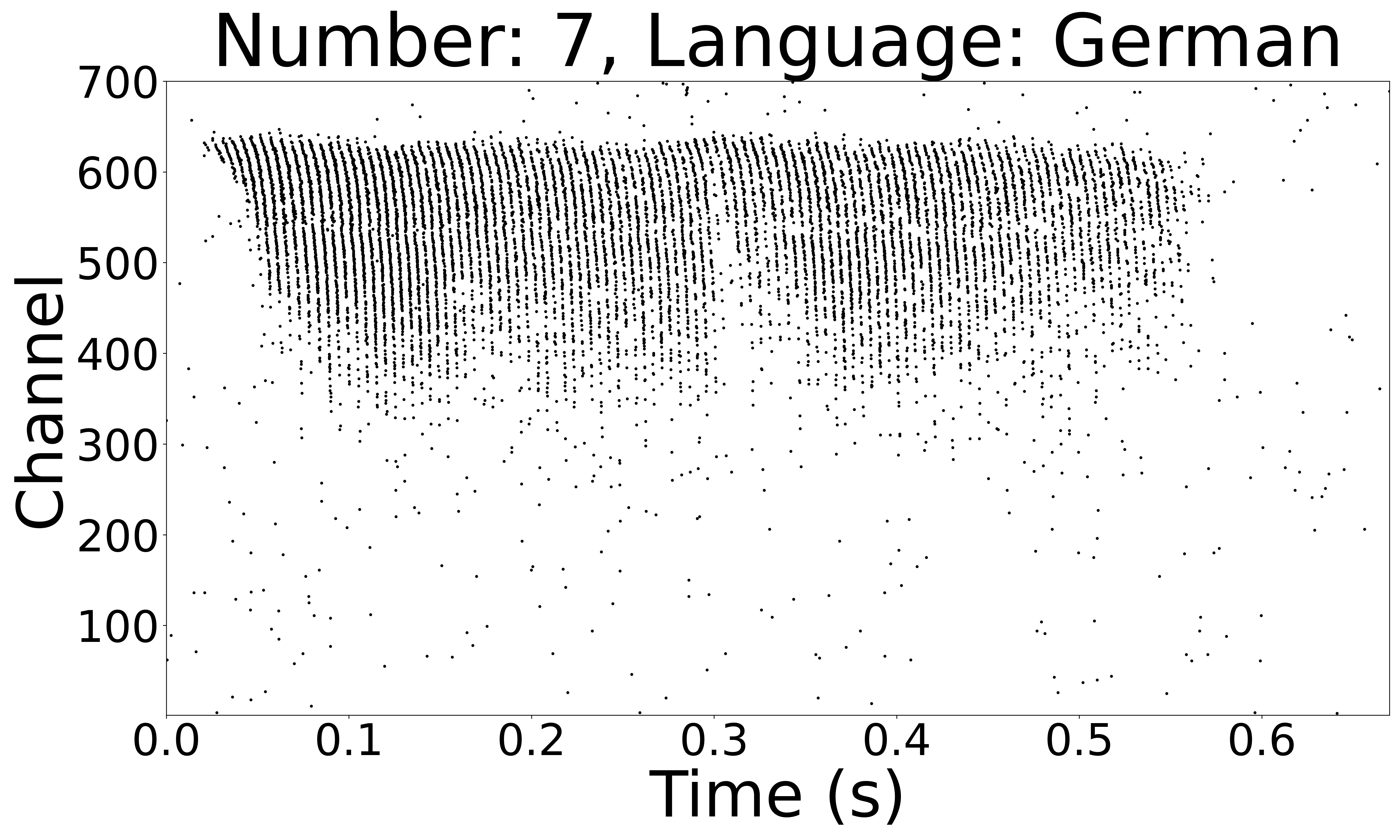}
    \end{minipage}

    \caption{Spectro-Temporal Spike Rasters from the SHD dataset.}
    \label{fig:visualisation}
\end{figure}

The advent of artificial cochlea models, especially dynamic audio sensors (DAS) \cite{liu2013asynchronous,mostafa2024}, has highlighted an opportunity to apply event-based AI methods, extensively explored in computer vision, for time-series applications. 
% DAS devices are inspired by biological auditory systems, enabling dynamic event output and efficient low-latency audio signal processing. This approach preserves fine-grained temporal information often lost in traditional frame-based methods, facilitating event-sequence processing.
A~popular benchmark for evaluating event-based time-series AI models is the Spiking Heidelberg Digits dataset \cite{cramer2020}. SHD simulates an AC by filtering recordings with a~computational model of the inner-ear. The dataset contains over 10k recordings (8156 train and 2264 test samples)  consisting of spoken digits in English and German from zero to nine. This results in a~set of sparse 700-channel spectrograms, each 750 ms in length on average (Figure \ref{fig:visualisation}).
In order to compare our SoC FPGA implementation of an event-graph neural network to previous state-of-the-art software models and hardware realisations, we use the SHD dataset with the same train/test split as a~benchmark in this paper.

% that processes audio data from a standard audio sensor into spike trains. These spike trains retain temporal information similar to the output of a DAS, making SHD a widely accepted dataset for benchmarking event-based algorithms. SHD consists of approximately 10,420 recordings of the digits “0” to “9” spoken in English and German, each lasting approximately 750 ms and consiting of 700 frequency channels. 
% Using temporal features, it serves as a valuable benchmark for evaluating event-based algorithms, including SNNs and event-based graph neural networks (GNNs). %Consequently, SHD effectively mimics the characteristics of data generated by DAS-like event-based sensors and provides a robust foundation for assessing our approach.

\subsection{Event-Based AI Applied to the SHD Dataset}

In Table \ref{tab:model_comparison} existing works applied to the SHD dataset are summarised. They mostly correspond to software implementations of recurrent spiking neural networks \cite{cramer2020,bittar2022,dampfhoffer2022,rossbroich2022} and feed-forward models leveraging learned synaptic delays \cite{dagostino2024,hammouamri2023,malettira2024,sun2023,yu2022}.

Recently, an alternative approach was proposed in \cite{lars} based on a~radically different idea. Instead of applying a~spiking neural network to raw data, an event-stream is first transformed into a~graph data structure, which is then processed using graph neural networks (GNNs). This event-graph neural network approach achieved a~performance close to the state-of-the-art synaptic delay-based SNNs on the SHD benchmark, although requiring between one and two orders of magnitude fewer synaptic weights in many cases. Notably, a~tiny model comprising only 17k parameters obtained a~test accuracy of over $90\%$. 

% % Among these, the study by \cite{lars} is particularly noteworthy.
% The study considers two configurations of the event graph model: a smaller version with a layer size of $p = 64$, comprising 17.1k synaptic weights, and a larger version with $p = 256$, containing 217k synaptic weights. Encouragingly, even the smaller event graph model demonstrates highly competitive results when compared to SNN methods. The larger model achieves a test accuracy of 94.3\%, only slightly below the reported state-of-the-art accuracy of 95.1\% \cite{hammouamri2023}. 

These findings suggest that the event-graph approach holds great promise for processing event-based time-series data. While it is believed that event-graphs, due to their deterministic synaptic weight access pattern and their natively asynchronous operation, may translate well into a~dedicated hardware, this has yet to be investigated.

\begin{table}[t!]
\centering
\caption{Model comparison on the SHD dataset. The symbol | indicates that only two models are present, while ``–'' denotes several intermediate values.}
\renewcommand{\arraystretch}{1.3}
\setlength{\tabcolsep}{4pt}
\resizebox{\textwidth}{!}{
\begin{tabular}{@{}lcccll@{}}
\toprule
\textbf{Author(s)} & \textbf{Year} & \textbf{Acc. (\%)} & \textbf{\#Params} & \textbf{Features/Details} \\ 
\midrule
%Cramer et al. \cite{cramer2020}        & 2020 & 48.1  & -      & Feed-forward (FF) SNN                     \\
Cramer et al. \cite{cramer2020}        & 2020 & 83.2  & -      & Recurrent SNN (RSNN)                      \\
%Perez-Nieves et al. \cite{perez2021}   & 2021 & 82.7  & -      & RSNN                                     \\
%Yao et al. \cite{yao2021}           & 2021 & 91.1  & -      & RSNN with attention                       \\
Yu et al.  \cite{yu2022}           & 2022 & 92.4  & 2.1M   & RSNN with synaptic dynamics              \\
Dampfhoffer et al. \cite{dampfhoffer2022}  & 2022 & 87.8  & 139k   & RSNN                                      \\
Bittar et al.     \cite{bittar2022}    & 2022 & 94.6  & 3.9M   & RSNN                                     \\
Rossbroich et al. \cite{rossbroich2022}   & 2022 & 83.5  & 209k   & Convolutional RSNN                       \\
%Hammouamri et al. \cite{hammouamri2023}   & 2023 & 93.4  & 2.7M   & RSNN with temporal convolution           \\
Hammouamri et al.  \cite{hammouamri2023}   & 2023 & 95.1  & 200k   & FF SNN with delay learning               \\
Sun et al.        \cite{sun2023}   & 2023 & 92.4  & 140k   & FF SNN with delay learning                \\
%Rafeldt et al. (Small) \cite{lars} & 2024 & 90.0  & 17.1k  & Small spectro-temporal graph             \\
%Rafeldt et al. (Large) \cite{lars}  & 2024 & 94.3  & 217k   & Large spectro-temporal graph            \\
Rafeldt et al. \cite{lars}  & 2024 & 90.0 | 94.3  & 17.1 | 217k   & Spectro-temporal graph            \\
D’Agostino et al. \cite{dagostino2024}   & 2024 & 87.6  & 224k   & FF SNN with dendritic delays              \\
Malettira et al.    \cite{malettira2024}   & 2024 & 94.7  & 1.3M   & Temporal Skips with delay learning      \\
%Karilanova et al.  \cite{karilanova2024}  & 2024 & 91.6  & -   & Zero-shot domain adaptation               \\ 
Carpegna et al.  \cite{carpegna2024spiker} & 2024 & 72.9 & - & RSNN on FPGA \\
Matinizadeh et al. \cite{quantisenc2024} & 2024 & 87.8 & - & Fully configurable FPGA SNN\\
The proposed system & 2025 & 88.8 - 94.5  & 8.6 - 272k   &  Spectro-temporal graph on SoC FPGA               \\ 
\bottomrule
\end{tabular}
}
\label{tab:model_comparison}
\end{table}

%\subsection{FPGA-based event-graphs implementations?}

%@Manon -> we have a 15 page limit and are at 12 without some important details ;) Yes but I think there is some redundancy (SHD described 3x) + maybe not doing a full section but for me this is the most important related work and it is described nowhere... / Yes. I think we could also drop some papers from the above table. - regarding the SOTA on this topic, we should have something ready.

\subsection{FPGA-Based Event-Sensor Data Processing}

%TODO Here a short paragraph on DAS = processing. We can start with my review and then concntrate on 2 papers - but also very shortly.

In contrast, numerous dedicated implementations of SNNs on application specific integrated circuits \cite{basu2022spiking} as well as FPGAs \cite{kryjak2024event} have been proposed. The majority of these works have focussed on event-vision applications. 
To the best of the authors' knowledge, there are three research papers describing FPGA-based SNN implementations for AC time-series processing \cite{carpegna2024spiker,quantisenc2024,xu2023event} -- only two of them report benchmarking results on the SHD dataset and are therefore included in Table \ref{tab:model_comparison}. The last article \cite{xu2023event} reports the results on custom-generated events from TIDIGITS audio dataset only, therefore it cannot be directly compared.

%The Spiker+ proposal \cite{spiker2024} introduces a framework aimed at providing low-power and efficient SNN accelerators for edge computing environments. It permits configurable multi-layer recurrent and feed-forward SNNs to be mapped onto an FPGA. Their evaluation on the SHD benchmark showed that Spiker+ consumes on average 430 mW of power with an inference latency of 0.54 ms while achieving a test accuracy of 72.9\%. 
%Similarly, QUANTISENC \cite{quantisenc2024}, an open-source tool for SNN quantisation and mapping to FPGA platforms. On the SHD dataset, a hardware implementation demonstrated an accuracy of 87.8\% and a reported peak power consumption of 1.6W. 

The Spiker+ framework \cite{carpegna2024spiker} and QUANTISENC tool \cite{quantisenc2024} both target efficient SNN accelerators for FPGA-based edge computing, with Spiker+ achieving 72.9\% accuracy on the SHD benchmark at 430 mW and 540 $\mu$s latency, while QUANTISENC demonstrated 87.8\% accuracy and 1.6 W peak power consumption. A~summary of these results is provided in Table \ref{tab:comparison}.

% \subsection{Summary}
% Based on the literature studies, we can draw several conclusions:
% %Existing literature has established:
% \begin{itemize}
%     \item SNNs can excel in capturing temporal dynamics in audio signals, but often come with large model sizes.
%     \item GCN-based spectro-temporal event-graphs offer parameter-efficient, high-accuracy solutions for event-driven audio recognition.
%     \item FPGAs can provide low-latency, energy-efficient acceleration for GCN-based AI models used for Dynamic Vision Sensor data processing. 
%     \item GCN-based audio event processing on FPGA remains an unexplored area.
% \end{itemize}

The main objective of this paper was to propose an FPGA event-graph implementation and compare the task accuracy, power consumption, latency and resource utilisation relative to these two papers on the SHD benchmark. This would provide an understanding of whether the promised performance of event-graph neural network can be achieved in hardware and how such an approach compares relative to SNNs. 
An additional goal of the implemented system was to enable fully asynchronous, event-by-event processing. %The hardware module accepts events with time intervals that depend on the difference in timestamps. 

% ur work addresses this gap by presenting the first FPGA-based implementation of a GCN-driven, spectro-temporal event-graph approach for audio recognition on the SHD dataset. Our solution demonstrates that hardware-aware optimisations, combined with efficient asynchronously updated event-based representations, can deliver competitive accuracy at low latency and power, making it well-suited for a wide range of edge applications.

% --------------------------------------------------------------------------

\section{FPGA Implementation of an Event-Graph for Time-Series Classification}
\label{sec:proposed_system}

%\begin{figure}
%    \centering
%    \includegraphics[width=0.5\linewidth]{gcn_audio.png}
%    \caption{Enter Caption}
%    \label{fig:enter-label}
%\end{figure}

%OLD

%The starting point for our work was the spectro-temporal graph described in the paper \cite{lars}, which was modified to take full advantage of the platform used, i.e. low latency and relative energy efficiency, while maintaining classification accuracy.
%The computational tasks were divided between the processing system (PS) and the programmable logic (PL) of the SoC FPGA device (cf. Figure \ref{}).
%In the PS we implemented the data input (SD card) and output (terminal) and the HEAD of the used neural network model, whilst PL was used to accelerate the feature generation backbone consisting of two submodules -- graph generation and graph convolutions. 

%NEW

The principle of the event-graph approach is to generate a~graph from the raw events generated by an event-sensor -- in this case the artificial cochlea.
Formally, a~graph is defined as $\mathcal{G} = (\mathcal{V},\mathcal{E})$, where $\mathcal{V}$ is the set of vertices (events) and $\mathcal{E}$ is the set of edges. Each vertex $v \in \mathcal{V}$ is associated with a~position $\mathcal{P}$ and feature vector $\mathcal{X}$, both of which characterise the underlying entity that the event represents, while edges represent the connections between pairs of these entities.

Spectro-temporal event-graphs are a~specific form of event-graphs constructed from AC time-series data. They are typically created by performing a~hemispherical search in the channel-time domain (Figure \ref{fig:visualisation}), establishing edges between events that lie within defined distance.
This neighbourhood can be determined by spatial and temporal radius thresholds, or, alternatively, edges can be assigned randomly within the search volume. Each new event generated by the sensor forms directed edges from previously recorded events found within a~semi-circle defined by a~channel radius $r_{ch}$ and a~time radius $r_t$.

In earlier work \cite{lars}, PointNetConv graph convolutions \cite{POINTNET} were applied to spectro-temporal event-graphs. In this publication, the feature vector \(\mathcal{X}\) of each event consisted of two components of a~normal vector estimated by fitting a~local surface to the event-data using a~least-squares approach. The position \(\mathcal{P}\) comprised a~channel index and a~timestamp. The relative positions between events defined edge vectors, creating an \(\mathcal{N}(i)\) neighbourhood set for each node \(i\).

Given an event-graph, we can apply PointNetConv operations across \(L\) convolutional layers. Each layer uses a~unique weight matrix shared across all events. Within each layer, the following operation is performed to update the embedding of the \(i^{th}\) event:

% \begin{equation}
% v_i^l = \phi_{agg}(\sigma(W^l \cdot [v_1^{l-1}; e_{i,1}]), \sigma(W^l \cdot [v_2^{l-1}; e_{i,2}]), \ldots, \sigma(W^l \cdot [v_M^{l-1}; e_{i,M}]))
% \end{equation}

\begin{equation}
\label{eq:pointnetconv}
\hat{\mathcal{X}}_i = \max_{j \in \mathcal{N}(i)} \bigl( \phi([\mathcal{X}_j \;||\; (\mathcal{P}_j - \mathcal{P}_i)]) \bigr),
\end{equation}

\noindent where \(\hat{\mathcal{X}}_i\) is the updated feature vector for event \(i\), and \(\phi(\cdot)\) represents a~fully-connected layer. The notation \(||\) indicates the concatenation of the feature vector \(\mathcal{X}_j\) with the relative position vector \(\mathcal{P}_j - \mathcal{P}_i\). After this transformation, a~feature-wise max-pooling aggregates the neighbour contributions into a~single output vector per event, and a~ReLU activation is applied to introduce nonlinearity.

% The aggregation function $\phi_{agg}()$ retains the maximum feature-wise vector elements, similar to max pooling in convolutional neural networks and each vertex forms a self-loop edge with itself.
To generate a~final prediction for an entire time-series, global average pooling is applied across all event embeddings from the last convolutional layer. The resulting vector is then fed into a~small multi-layer perceptron, which produces the desired output (e.g. a~class prediction) for the time-series under consideration.

% The aim of this study was to develop a system capable of processing incoming events individually using spectro-temporal graph neural networks. The proposed model processes each event by establishing edges between neighbouring nodes to form a graph, passing it through graph convolutional layers, and generating a representative vector of the processed data, which can subsequently be used by a classifier. This approach is based on the work presented in \cite{lars}, which we have adapted and modified to take full advantage of hardware acceleration.

The following subsections detail the adaptations and optimisations performed in order to map this software baseline \cite{lars} to a~hardware accelerator architecture on FPGA.
To achieve more flexible and scalable implementation, computational tasks were distributed between the processing system (PS) and the programmable logic (PL) of the SoC FPGA device, as illustrated in Figure \ref{fig:teaser}.
\subsection{Graph Generator}
%\begin{figure}
%    \centering
%    \includegraphics[width=1\linewidth]{graph--gen--v2.png}
%    \caption{Overview of the graph generation module. In this figure, the skip step is set to 10 and \( r_{ch} \) is set to 100.}
%    \label{fig:graph--gen-v2}
%\end{figure}
\begin{figure}
    \centering
    \includegraphics[width=1\linewidth]{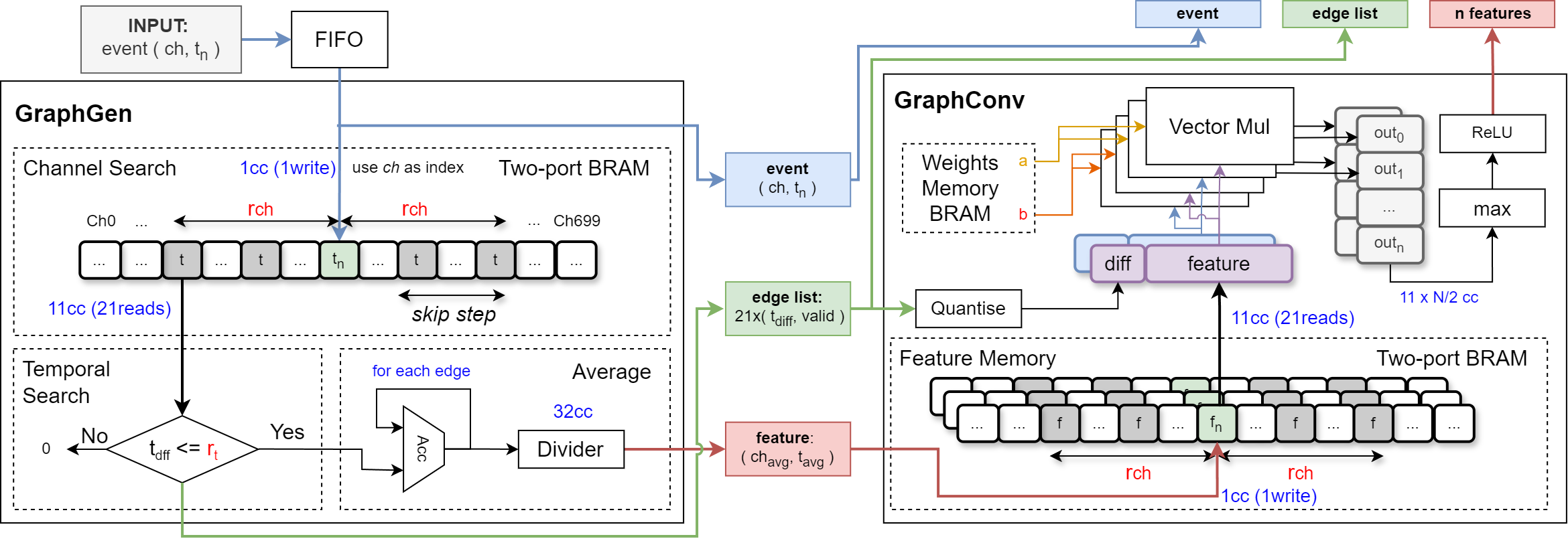}
    \caption{Overview of the graph generation and convolution modules. The \textit{skip step} is set to 10 and \( r_{ch} \) is set to 100. The number of output features is denoted as $n$.}
    \label{fig:graph--gen-v2}
\end{figure}

%In the baseline model, each event is represented by its position, determined by the timestamp \( t \) and the channel index \( f \). Additionally the features of a normal vector were calculated for each event from its local neighbourhood. Furthermore, directed edges were constructed using a hemispherical search algorithm inspired by \cite{HUGNET}, selecting up to \( M \) events within predefined spectro-temporal radii (time \( r_t \), channel \( r_{ch} \)).

The baseline model presented in \cite{lars} has proven effective in capturing spectro-temporal relationships within event-graphs. However, it faces significant challenges when considering hardware implementation. First, the calculation of a~normal vector requires fitting regression lines to the event-data that has a~high computational complexity. Second, identifying neighbouring vertices requires storing all vertex vectors in memory and performing sequential searches, resulting in high latency and significant memory overhead.

To address these limitations, a~novel graph generation was designed to optimise memory usage, reduce latency, and minimise computational overhead. The following key modifications were introduced:

\begin{enumerate}
    \item Drawing inspiration from FPGA-based event camera data processing implementations \cite{jeziorek2024embedded,yang2024evgnn}, events are stored in 1D context memory (implemented as a~block RAM memory (BRAM)) using their channel \( ch \) as the address and timestamps \( t \) as the data. Only the most recent event generated per channel is stored, continuously overwriting timestamps at each channel index. Each new event can be connected only to the ones already processed, creating a~directed graph. 
    
    \item Event-normal vectors were replaced with simpler features based on the average timestamp and channel coordinates of neighbouring events, massively reducing the computational cost.

    \item To improve the efficiency of neighbour search, we introduce a~method called \textit{skip step connection}. This corresponds to a~pre-defined deterministic pattern regarding how edges can be formed between vertices.
\end{enumerate}

% This graph generation strategy is well-suited for hardware, as it requires only a limited number of stored values. By c, the graph remains temporally aligned, supporting asynchronous event processing. 

% Surprisingly, despite the large simplification, studies using our event-graph software model showed that using the average position of all neighbours yields almost equivalent performance relative to using the normal vector. Additionally, as confirmed in Section~\ref{subsec:ablation}, the \textit{skip step} strategy significantly decreases computational complexity while preserving broad frequency coverage.

Figure~\ref{fig:graph--gen-v2} illustrates the main components and data flow of the graph generation module. Asynchronous events, defined by their time \( t \) and channel index \( ch \), enter a~first-in first-out (FIFO) buffer, ensuring a~stable data stream. Each event then moves to a~1D context memory indexed by the channel.

The module performs a~neighbour search using the \textit{skip step connection} method. Instead of scanning all possible neighbours, the system reads events at fixed intervals along the channel dimension from the BRAM. Events that meet the \textit{temporal search} criteria (based on \( r_t \) radius) are added to an edge list, and their corresponding time and channel indices are fed into an accumulator. After iterating over all channels, this accumulator computes the average position of the neighbours using a~simple divider.
The resulting averaged features, together with the newly stored event and the edge list, are then passed to the first stage of the graph convolution pipeline, where the time in channel index (\( ch)\) is updated.

The total number of clock cycles \( N_{\text{cycles}} \) required for graph generation depends on the channel radius (\( r_{ch} \)), the \textit{skip step} (\( s \)), and the additional cycles for feature computation (\( N_{\text{div}} \)). Assuming dual-port memory (two reads per clock cycle), this can be expressed as:
\begin{equation}
N_{\text{cycles}} = \frac{1 + 2 \cdot \frac{r_{ch}}{s}}{2} + N_{\text{div}}
\end{equation}
Here, \(1\) corresponds to the central channel read, \(2 \cdot \frac{r_{ch}}{s}\) represents the reads from the upper and lower channels, and \( N_{\text{div}} \) accounts for division.

% ----------------------------------------------------------------------------------------------------
\subsection{Graph Convolution}

In the baseline approach, the PointNetConv \cite{POINTNET} convolutional layer was used, which extends the classical \textit{message passing mechanism} commonly used in graph convolutions. 
As highlighted in \cite{jeziorek2023memory,jeziorek2024embedded}, these layers are lightweight and well-suited for hardware acceleration, making them a~natural choice for adoption in our work. However, we introduce two key modifications.

% Consistent with the baseline method, we omit the \(\gamma\) function and define \(\phi\) as a simple linear function.  

First, we integrate a~\textit{batch normalisation (BN)} layer into \(\phi\) to ensure stable training. Notably, during quantisation these layers are folded \cite{jacob2018quantization}, which prevents any increase in the model parameters for hardware deployment. The second and more important change involves an additional normalisation step for the positional differences used on the event-graph edges. Input data is initially normalised to \((0, 1)\) range, while neighbours are determined within the radius \(r_{ch}\) and \(r_t\). Consequently, channel index differences are within \((-r_{ch}, r_{ch})\), and time index differences are within \((-r_t, 0)\) due to the use of a~time-based directed graph generator. These values occupy only a~small fraction of \((0, 1)\) range, which adversely affects training and reduces precision during quantisation. To address this, we apply \textit{positional normalisation (PN)} after computing position differences, rescaling the values back to \((0, 1)\) range, by multiplying time differences by \(-\frac{1}{r_t}\), and channel differences by adding \(r_{ch}\) and multiplying by \(\frac{2}{r_{ch}}\).

The modified model is expressed as:

\begin{equation}
\label{eq:pointnetconv}
\hat{\mathcal{X}_i} = \max_{j \in N(i)} (BN\ \phi([\mathcal{X}_j || PN(\mathcal{P}_j - \mathcal{P}_i)])).
\end{equation}

For hardware implementation each graph convolution assumes fully asynchronous event-by-event processing and can be executed in parallel.
Each incoming event with its edge list and input features is processed independently. 
The module must also access the features of each vertex connected with an edge and the difference between the neighbour and the processed event.
For this purpose, we implemented one 2-port BRAM memory that stores the features of the last processed event for each channel, and another one for storing the weights. 

To determine the final output feature, the linear layer must be applied for the vertex itself (so-called self-loop) and for each of its neighbours (a~maximum of \texttt{MAX\_EDGE} = 21 times with the \(r_{ch}\)=100 and \textit{skip step} \(s\)=10). 
The key part of the graph convolution is the vector multiplication module -- in order to reduce resource utilisation, the module was implemented in a~way that some of the calculations were performed sequentially (as inspired by \cite{jeziorek2024embedded}). 

Taking advantage of two-port memories, we process two feature vectors and calculate two elements of the output vector (\texttt{OUT\_DIM} = 64) at the same time (cf. Figure \ref{fig:graph--gen-v2}). The total number of clock cycles required to perform the convolution for a~single event can be determined with the following formula:

\begin{equation}
\label{eq:pointnetconvHW}
N_{\text{cycles}} = \frac{MAX\_EDGE+1}{2}\cdot \frac{OUT\_DIM}{2} 
\end{equation}
 
The graph convolution module can be considered a~bottleneck of the proposed method. The throughput of the system is dependent on the maximum number of clock cycles required per single event. For our baseline model implemented for a~200 MHz clock, the theoretical throughput was thus calculated to be 555 kEPS (thousand events per second). At the same time, the average in the SHD dataset is around 20 kEPS. 
However, it should be noted that the number of parallel vector multiplication modules could be increased to improve latency and throughput (using additional resources), or decreased with an opposite effect.
%We plan to investigate the dependence of resource consumption on latency in future studies. 

An essential part of the system is the quantisation. In our work, the precision of calculations on integer values can be selected per layer. 
Both feature map elements and weights are stored in memory as unsigned integers, and rescaled before (quantisation) and after (re-quantisation) of multiplications. 
For scaling we use DSP multiplication and bit-shifting or look-up tables (depending on the number of possible quantised values). 
%Based on the experiments performed, we determined that the satisfactory size/accuracy ratio can be achieved with 16 bits of precision for the first convolutional layer, and only 8 bits for each subsequent layer. 

%----------------------------------------------------------------------------------------------------
\subsection{Graph Average Pool}
%TK - OK

The final step of the baseline method requires global average pooling to aggregate all of the vertex-level features of the last layer into a~single fixed-length vector to be processed by fully-connected layers. Our experiments with simpler alternatives, such as global maximum pooling or global sum pooling, revealed that these approaches result in significantly degraded quality results (by 7 pps for max and 5 pps for sum). Therefore, we decided to keep the original method.

%\todo{HERE HW IMPLEMENTATION}
We implemented a~global average pooling module which receives the features from the last stage of the graph convolution (Figure \ref{fig:teaser}) to the input. % along with a ``valid'' flag. A further output indicates whether the event is the final event before a prediction should be made regarding the processed event-graph. 
%Note that this is specific for the application to the SHD dataset since the beginning and end of spoken-words are defined. For streaming time-series data this signal could be set high, for example, with a user-defined periodicity. 
The mean value is calculated by accumulating the sum of each vector feature in a~given register. %, sized to prevent overflow, when the valid flag is set. For example, if the output vector is a 64-size vector, 64 accumulator registers are needed. 
To determine the number of events that have been accumulated, a~simple counter is used.
When the output vector corresponding to the last event in a~spoken word is calculated, the accumulator register values are divided by the current value of the counter. These values are then output by the module and stored in a~BRAM memory. %The final part of the Full Connected Layers are performed by the SoC FPGA processor.
% Therefore, we receive the 64-element average vector in a serial manner in 8-bit format, which along with the ``valid'' flag and the index of the element from the vector is passed to the output of the module.

%The final stage of the proposed method employs global average pooling to aggregate vertex-level feature representations into a single fixed-length vector, which is then used for classification by fully connected layers executed on the PS. The module accumulates feature vectors from all vertices and calculates the average by dividing the sum by the total number of vertices upon completion of processing.
% --------------------------------------------------------------------------

\subsection{System Integration}
%TK - OK

%MAYBE SMALL SECTION ABOUT DIVIDING TASKS BETWEEN PS AND PL, THAT WE SIMULATE EVENTS BY SD CARD ECT..

To test the architecture on the target platform, it was necessary to provide the data from an AC to the input and receive the output features to classify it.
As we currently do not have the access to the real AC sensor, we decided to use test data samples from the SHD dataset written into separate text files on an SD card, which were read using the SoC FPGA processing system.
The subsequent events were sent to the programmable logic via an AXI4 bus.
To simulate AC, it was required to provide events to the architecture at the moments that respect the event timestamps. In the processing system a~delay function was implemented in order to achieve this based on the differences of successive timestamps.

When the average pooling output is fully written into the BRAM memory, an interrupt flag is raised, which triggers a~function in the SoC FPGA processing system. The BRAM contents are read into the PS and two multi-layer perceptron functions are executed. Each layer was implemented using a~pair of nested loops and is highly sequential. For a~future work, these calculations could also be realised with a~greater parallelism in the programmable logic. 
% We decided to implement the final linear layer on the PS side to make the solution more versatile, which then simply takes the maximum value, equivalent to the predicted class index of the data sample.

% We introduced achieved scores for software model (training, quantization, results, comparison with SOTA).
% We we describe in detail conducted ablation studies (already tested multiple methods of hardware-friendly graph generation [different R, skip-search], different features for first convolitions [Poly vs average frequency], features for network's HEAD [GlobalMaxPool, GlobalAveragePool, LSTM, maybe WeightedGlobalAveragePool]).

% We also introduce hardware-implementation, evaluate utilisation, timing, latency, throughput.
% We study the impact of the number of parallel calculations on latency and resource consumption. 
% We highlight the low-latency (output a couple of us after the last event). 
% It would be best to compare the power consumption for FPGA (Vivado-estimated), CPU and GPU (Jetson?). 

% Comparison with SOTA
% + ASIC (if possible) rough.

% --------------------------------------------------------------------------------------------------------------------------------------------------------
\section{Evaluation}
\label{sec:evaluation}

%In order to evaluate the event-graph model, we consider three key metrics: top-1 accuracy, model size and FLOPs per event on the SHD dataset \cite{cramer2020}. In order to dimension the model that will be mapped onto the hardware, we conducted preliminary ablation studies concerning graph generation and graph convolution hyper-parameters. This permits the runtime performance, power consumption and resource utilisation of the hardware implemented model to be compared to previous SNN FPGA implementations. 

In order to evaluate the event-graph model, we consider three key metrics: top-1 accuracy, model size and FLOPs per event on the SHD dataset \cite{cramer2020}. This permits the runtime performance, power consumption and resource utilisation of the hardware implemented model to be compared with previous SNN FPGA implementations. We conducted preliminary ablation studies concerning graph generation and graph convolution hyper-parameters to evaluate several model configurations and select the best one for the implementation in hardware.

\subsection{Setup}
\label{subsec:setup}

For implementation and training, we used PyTorch and PyTorch Geometric libraries \cite{Fey/Lenssen/2019}. All prepared models were trained for 100 epochs in floating-point precision, followed by 20 epochs of quantisation-aware training \cite{jacob2018quantization} with a~batch size of 16 on a~single A100 GPU. The Adam optimiser was configured with a~learning rate of 2e-4, weight decay of 1e-4, and a~ReduceLROnPlateau scheduler with a~0.5 reduction factor and patience of 10. The best model weights were saved based on the minimum training loss and subsequently evaluated on the test set.

All models consist of four PointNetConv convolutional layers, a~graph average pooling layer, and two fully-connected layers. To preserve the input temporal resolution, we applied 16-bit quantisation on the first convolution. The remaining convolutional layers were quantised with 8-bit precision. The final two fully-connected layers, implemented in the processing system of the SoC FPGA device, were not quantised and therefore use 32-bit floating-point.

The hardware architecture of the network was implemented in the SystemVerilog language using Vivado 2022.2, while the processing system was programmed using Vitis 2022.2 in the C++ language.
After verifying the compatibility with PyTorch software model via Vivado simulation, it was implemented on a~Zynq UltraScale+ ZCU104 FPGA platform with XCZU7EV chip from AMD Xilinx.

\subsection{Ablation Studies}
\label{subsec:ablation}

In this section, we present the results of experiments designed to evaluate the impact of hyper-parameters on the proposed architecture. 
First, we study the impact of the graph generator parameters. We then compare the performance and number of parameters required for different variants of the model in order to evaluate the scalability of the proposed approach.
Unless stated otherwise, the base model was configured with all output channels set to 64, a~graph generator with an \(r_{ch}\) of 100, a~\textit{skip step} of 10 and an \(r_t\) of 20 ms.

\subsubsection{Graph Generation Configurations}

In Table \ref{tab:graph_gen} we present the results of our accuracy analysis for the base model as a~function of the \(r_{ch}\) and \textit{skip step} parameters. The experiment was carried out in two stages. In the first one, we varied the \(r_{ch}\) parameter from 30 to 300 while keeping the \textit{skip step} fixed at 10. In the second stage, using the best \(r_{ch}\) value identified in the first stage, we varied the \textit{skip step} parameter from 1 to 20. A~``–''~score indicates configurations excluded from the analysis due to excessively large tensor sizes that could not fit in the GPU memory.

\begin{table}[!h]
\caption{The impact of search radius \(r_{ch}\) and \textit{skip step} parameters on top-1 accuracy. The best results, highlighted in \textbf{bold}, were obtained for moderate parameters.}
\resizebox{\textwidth}{!}{%
\begin{tabular}{@{}cccc|cccc@{}}
\toprule
~~\(r_{ch}\)~~ &
  ~~\textit{skip step}~~ &
  \begin{tabular}[c]{@{}c@{}}~~top-1 acc.~~\\ float\end{tabular} &
  \begin{tabular}[c]{@{}c@{}}~~top-1 acc.~~\\ quantised\end{tabular} &
  ~~\(r_{ch}\)~~ &
  ~~\textit{skip step}~~ &
  \begin{tabular}[c]{@{}c@{}}~~top-1 acc.~~\\ float\end{tabular} &
  \begin{tabular}[c]{@{}c@{}}~~top-1 acc.~~\\ quantised\end{tabular} \\ \midrule
100 & 1  & 90.95 \% & -        & 30  & 10 & 85.43 \% & 84.23 \% \\
100 & 5  & 91.37 \% & 90.10 \% & 50  & 10 & 88.25 \% & 87.60 \% \\
\textbf{100} & \textbf{10} & \textbf{92.74 \%} & \textbf{92.30 \%} & 200 & 10 & 91.23 \% & 90.05 \% \\
100 & 15 & 90.63 \% & 89.22 \% & 250 & 10 & 90.47 \% & 89.72 \% \\
100 & 20 & 90.05 \% & 88.91 \% & 300 & 10 & 89.66 \% & 88.62 \% \\ \bottomrule
\end{tabular}%
\label{tab:graph_gen}
}
\end{table}

We observed that too few edges -- caused by either a~high \textit{skip step} or a~low \(r_{ch}\) -- negatively affect the model accuracy. Conversely, an excessively large \(r_{ch}\) or too small \textit{skip step} can also degrade classification performance. This is due to the sensitivity of PointNetConv layers to outliers, which become more noticeable with a~large number of edges -- a~consequence of using a~max-type aggregation. The best results were achieved with moderate parameter values: an \(r_{ch}\) of 100 and a~\textit{skip step} of 10, yielding 92.74\% accuracy for the floating-point model. %Although it is possible that simultaneously increasing both the \(r_{ch}\) and the \textit{skip step} might lead to even better performance, the chosen parameters were sufficient for our study.

\begin{figure}[ht]
    \centering
    \begin{subfigure}[b]{0.49\textwidth}
        \centering
        \includegraphics[width=\textwidth]{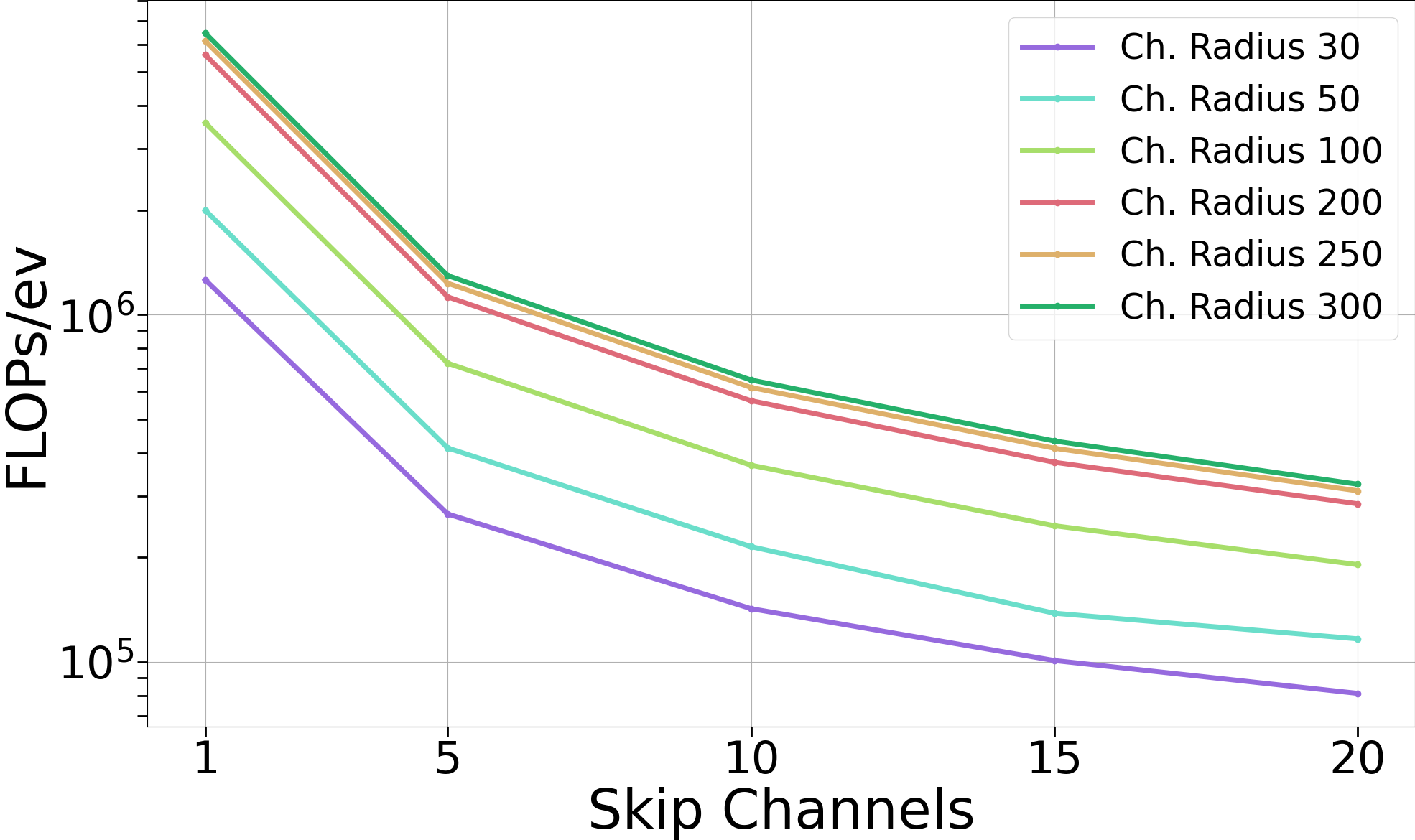}
        \caption{Average FLOPs per event for different graph generation configurations.}
        \label{fig:subfig1}
    \end{subfigure}
    \hfill
    \begin{subfigure}[b]{0.49\textwidth}
        \centering
        \includegraphics[width=\textwidth]{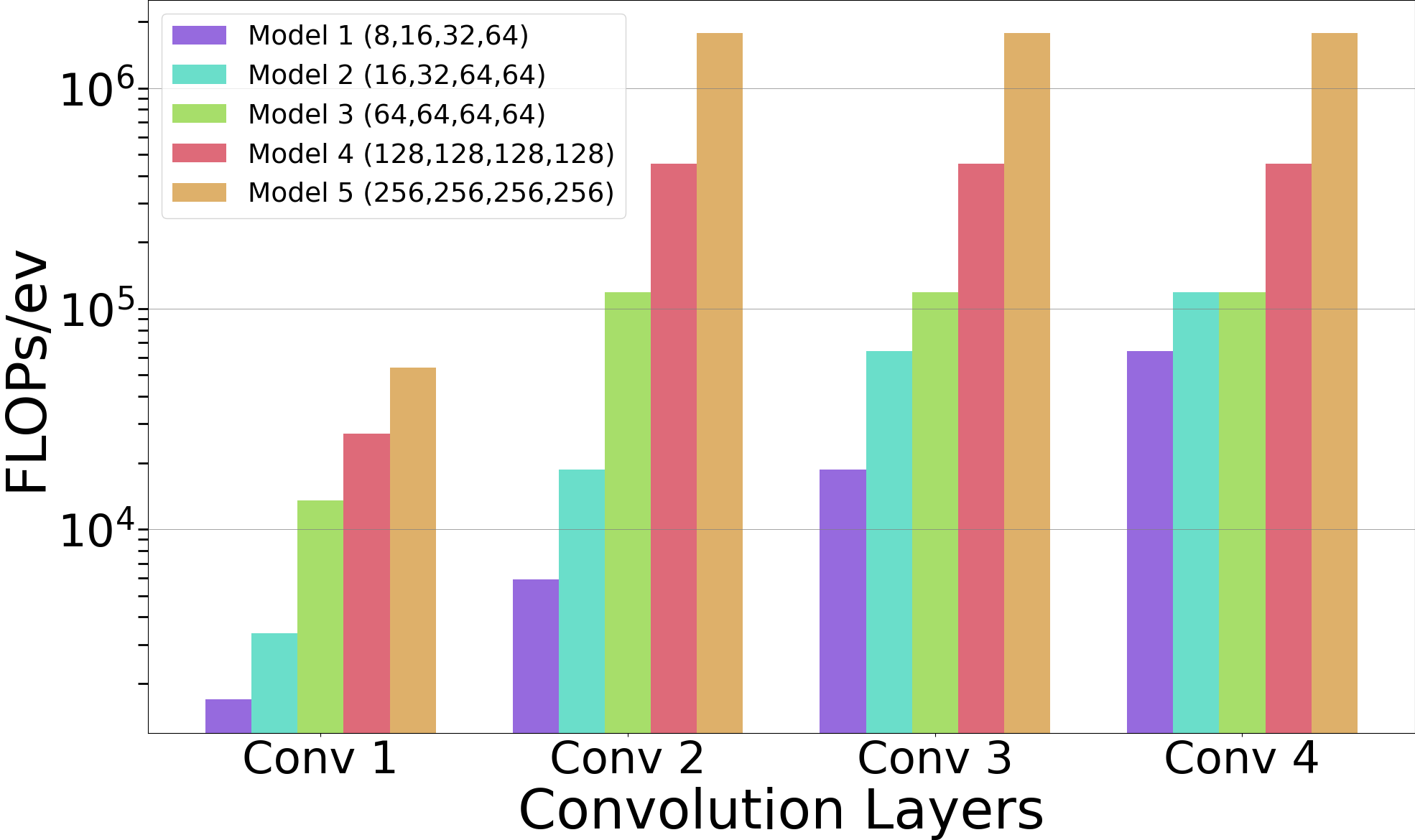}
        \caption{Average FLOPs per event for model configurations (outputs per convolution).}
        \label{fig:subfig2}
    \end{subfigure}
    \caption{Model complexity analysis in term of FLOPs. The figures illustrate not only the significant impact of model size on computational complexity, but also the parameters of the graph generator.}
    \label{fig:two-plots}
\end{figure}

It is also important to emphasise that the computational cost of graph convolutions is proportional to the number of edges. As illustrated in Figure \ref{fig:subfig1}, the number of floating-point operations (FLOPs) required per event -- for all convolutions, averaged across the entire test dataset -- is influenced by the \textit{skip step} parameter. This analysis indicates that using connections with a~\textit{skip step} of 10 reduces FLOPs by a~factor of ten (3.568 vs 0.368 MFLOPs/ev), confirming a~linear relationship between computational demands and the number of generated edges. The role of our novel \textit{skip step} method is crucial for reducing the calculation complexity. Note that this is not specific to the context of FPGA implementations.

\subsubsection{Model Configurations}

%\vspace{-1cm}

In this subsection, we examine how varying the dimensions of the convolutional layers affects model performance. As a~reference, we used two model configurations described in \cite{lars}: small -- with all layer dimensions set to 64 and large -- with all layer dimensions set to 256. To explore a~broader range of options, we study three additional models: one with all layer sizes set to 128, and two variants with increasing layer size. The results, including layer dimensions, total number of parameters, and classification accuracy, are summarised in Table \ref{tab:model_size}.

The results demonstrate that increasing the number of parameters improves classification accuracy, highlighting the scalability of our method. Importantly, compared to the baseline (achieving 90.0\% and 94.3\% accuracy for the small and large models), the equivalent models based on our proposed approach (minor differences in parameter number are due to BatchNorm layers) achieve accuracies of 92.74\% and 94.63\%, respectively. This indicates that the proposed solutions not only facilitate efficient hardware implementation, but also enhance performance.

\begin{table}[!h]
\centering
\caption{Layer configurations (number of output features) with number of parameters and top-1 accuracy. Increasing the number of parameters improves classification results, demonstrating the scalability of our approach.}
\resizebox{\textwidth}{!}{%
\begin{tabular}{@{}cccccccc@{}}
\toprule
  ~~Conv 1~~ &
  ~~Conv 2~~ &
  ~~Conv 3~~ &
  ~~Conv 4~~ &
  ~~FC~~ &
  ~~Params (k)~~ &
  \begin{tabular}[c]{@{}c@{}}~~top-1 acc.~~\\ float\end{tabular} &
  \begin{tabular}[c]{@{}c@{}}~~top-1 acc.~~\\ ~quantised\end{tabular} \\ \midrule
8   & 16  & 32  & 64  & 64  & 8.6   & 89.22 \% & 88.78 \% \\
16  & 32  & 64  & 64  & 64  & 12.9  & 90.54 \% & 90.98 \% \\
64  & 64  & 64  & 64  & 64  & 18.9  & 92.74 \% & 92.30 \% \\
128 & 128 & 128 & 128 & 128 & 70.5  & 93.93 \% & 93.31 \% \\
256 & 256 & 256 & 256 & 256 & 272.0 & 94.64 \% & 94.45 \% \\ \bottomrule
\end{tabular}%
}
\label{tab:model_size}
\end{table}

Similar to the number of edges, another key factor influencing model complexity is the dimensionality of event feature vectors processed by the network. To address this, we conducted an analysis of the average number of FLOPs per event for each convolutional layer, depending on its configuration. The results are presented in Figure \ref{fig:subfig2}. They show that the computational complexity of the first convolutional layer increases linearly with its dimension due to the fixed size of input data (13.516 vs 27.032 kFLOPs/ev for Model 3 vs Model 4), while for subsequent layers in models with uniform dimensions, complexity grows quadratically (118.269 vs 452.801 kFLOPs/ev for Conv 2). Consequently, the total FLOPs for the evaluated models are 0.091, 0.204, 0.368, 1.385, and 5.366 MFLOPs/ev, respectively. These results underline the importance of careful selection of network parameters to optimise memory and computational costs.

\subsection{Comparison with the State-of-the-Art}

%This section compares our solution with methods from the literature that have been evaluated on the SHD dataset. 
In order to position our work relative to the state-of-the-art, we first compare accuracy and parameter count of our event-graph neural network software model to other event-based AI approaches on the SHD dataset. Next, we compare our hardware implementation with two FPGA works of spiking neural networks with consideration of power consumption, latency and resource utilisation.

The results in Table \ref{tab:model_comparison} show that our models are among the smallest in the state-of-the-art. Our base model, with only 18.9k parameters, in terms of size is surpassed only by the small model from \cite{lars}, which served as our baseline inspiration, and by some optimised variants of our network (from Table \ref{tab:model_size}). The test accuracy of 92.31\% is comparable to those reported by \cite{sun2023} and \cite{yu2022}, while drastically reducing the number of parameters by 86.5\% and over 99\%, respectively. Our largest model achieves an accuracy of 94.45\%. This result is lower than only three models. It should be noted that two of these models \cite{bittar2022,malettira2024} require millions of parameters compared to only 272k in our case. 
% Only one solution \cite{hammouamri2023} outperforms ours by only 0.6\% while using 70k fewer parameters. 
Furthermore, it should be noted that our event-graph approach underwent significant hardware-aware modifications and quantisation.
%, which inevitably restricts their maximum achievable performance.
The high performance achieved is a~testament to the suitability of GNNs for event-data processing.

A~direct comparison of our hardware implementation with solutions from the literature (SNNs) is presented in Table \ref{tab:comparison}.
We implemented base and tiny variants (the latter refers to the smallest model in Table \ref{tab:model_size}) for Zynq US+ ZCU104 board.
%We verified the module's compatibility with the software model through both behavioural simulation and hardware testing.
%Moreover, 
%We evaluated system latency, resource utilisation and power consumption.
% 

\begin{table}[ht]
\centering
\caption{Comparison of state-of-the-art hardware implementations. The values in parentheses take into account the PS part (classification head). 
%LUT -- look-up table, FF -- flip-flop, BRAM -- block RAM memory, DSP -- digital signal processing block. 
The values marked with ``-'' were not mentioned in the respective papers.
%The resource utilisation for \cite{quantisenc2024} was reported only in terms of percentages; thus, the absolute values were estimated based on the maximum resources available on the ZCU104 board.
}

\resizebox{\textwidth}{!}{%
\begin{tabular}{@{}lcc|cc@{}}
\toprule
\textbf{}            & \textbf{Matinizadeh} \cite{quantisenc2024} & \textbf{Carpegna} \cite{carpegna2024spiker} &  \textbf{Ours (base)} & \textbf{Ours (tiny)}  \\\midrule
Device/Board       & Zynq ZCU104          & Zynq Z7-20              & Zynq ZCU104 & Zynq ZCU104 \\ %\multicolumn{2}{c}{Zynq ZCU104} \\
Logic cells        & -                      & 18,268               & - & - \\
LUT                & 149,760                  & -                    & 81,567 & 34,474 \\
FF                 & 92,160                   & -                    & 47,699 & 23,713 \\
BRAM               & 75                       & 51                   & 70 & 28 \\
DSP               & -                       & -                   & 318 & 106 \\
Frequency          & 100 MHz                       & 100 MHz             & 200 MHz & 200 MHz \\ %\multicolumn{2}{c}{200 MHz}  \\
Latency [$\mu$s]            & -                        & 540              & 8  (179)  & 4 (175)\\
Power [W]              & -                        & 0.43               & 1.20 (3.94) & 1.05 (3.79) \\
Peak power [W] & 1.629                  & -                    & - & - \\
Accuracy           & 87.80 \%                   & 72.99 \%             & \textbf{92.30 \%}  & 88.78 \% \\\bottomrule
\end{tabular}}
\label{tab:comparison}
\end{table}

With our approach, we established a~new state-of-the-art accuracy for an FPGA solution applied to the SHD benchmark by a~considerable 4.5\% margin. 
Our base model uses available logic resources more evenly than \cite{quantisenc2024} (fewer LUTs/FFs, more DSPs), thus avoiding potential routing delays. %is more efficient in terms of LUT/FF utilisation than \cite{quantisenc2024} while outperforming it by 4.5\% accuracy. 
We achieved this through executing some of the multiplications sequentially and leveraging the DSP multipliers.
Furthermore, with the tiny model we achieved 0.8\% higher accuracy with just 23\% of LUTs and 37\% of BRAMs compared to \cite{quantisenc2024}.

An important advantage of our solution is its asynchronous event-by-event processing and the intrinsic ability of event-graphs to exploit temporal sparsity for rapid and efficient calculations.
%Notably, our approach has the unique advantage of being able to process data immediately, as soon as an event is generated by the sensor. 
However, this makes it difficult to compare it with other solutions in terms of latency. 
Our base hardware module's (graph generation and graph convolution) per-event latency was measured at 8 $\mu$s for a~200 MHz clock with 555 kEPS throughput (4 $\mu$s and 277 kEPS for tiny model).
After taking into account the PS-PL communication and the determination of the value of classification results by the network's head in the processing part of the heterogeneous system, we obtain the predicted class 179 $\mu$s after the occurrence of the last event.
In \cite{carpegna2024spiker} the reported latency of 0.54 ms assumes the processing of the entire data sample after its encoding in 100 time steps. 
Our solution, which requires no prior data aggregation, maintains high throughput and low latency for edge applications by exploiting data's temporal sparsity.

The power estimation provided by the Vivado software for the PL part of the system (base model) is 1.20 W (0.60 W dynamic and 0.60 W static). With the PS part included energy usage increases to 3.94 W (3.25 W dynamic and 0.7 W static) for the entire base architecture.
%However, it is important to note that the power estimation tool does not take into account the substantial data sparsity. %-- in real case the actual average power consumption would be smaller, since the computations are performed only for processed events and active edges. Therefore, the logic elements are used only in case of new input data, which is different from synchronous processing (and calculated by the power estimation tool).

For an additional ablation study we implemented the resource-optimised variant of the base model for decreased number of parallel multiplications (from 4 per convolution to 2). 
With this modification, we were able to achieve 49\% decrease in both DSP and LUT utilisation and 35\% for FFs with a~simultaneous increase of latency for a~single event (in the PL) to 15.2 $\mu$s (theoretical throughput of 277 kEPS).
This experiment confirms that the amount of resources depends on the latency -- the choice of the final solution should be based on the requirements of the specific task and the size of the target platform.

\section{Summary}

In this work we present the first hardware-software co-design of an event-graph neural network for time-series audio classification. Because our approach exploits the inherent sparsity of event-data to reduce the computational complexity and latency, it is highly promising for near-sensor AI processing at the edge.

We proposed adaptations and optimisations permitting an event-graph to be implemented efficiently on a~SoC FPGA. In particular, we presented the novel \textit{skip step} graph generation method with simplified and computationally efficient features. In spite of this, our quantised hardware-aware event-graph model achieved a~test accuracy extremely close to floating-point precision software models from the state-of-the-art while requiring almost two orders of magnitude fewer parameters. Crucially, our method outperformed all previous FPGA implementations of hardware-aware spiking neural networks on the same benchmark  achieving improvements of up to 4.5\% and 19.3\% in accuracy. Relative to these works, the utilisation of FPGA resources and the latency were also reduced.
Our smallest model (tiny) achieves comparable accuracy to previous state-of-the-art while utilising only 23\% of logic and 37\% of memory resources.

We have confirmed that the hardware implementation of graph neural networks applied to event data from artificial cochlea sensors is both highly efficient and capable of achieving superior task accuracy compared to spiking neural network alternatives. 

In future work we intend to improve our system by making it compatible with continuously streamed events. Currently, we classify the samples, in which only one word is spoken. However, in real-life case there may be multiple words in a~recorded sentence that need to be separated and recognised, thus making this task more difficult.
%rely on knowing the beginning and end of a~spoken word, which is not realistic in real edge AI applications. 
Additionally, while input for our system was simulated using event-data written to an SD card, our goal is to integrate a~real event-sensor for a~fully end-to-end demonstrator. This may not necessarily be an audio sensor, our approach is generally applicable to any time-series application. Furthermore, in the current version of the system the multi-layer perceptron that outputs the final prediction result is implemented in the processing system of the used SoC FPGA for a~higher flexibility and scalability of the solution. Not only we intend to perform this function in the programmable logic or in the AI cores available in the Versal SoC FPGA for a~greater parallelism and lower latency, but we also plan to explore different neural network architectures that may be better adapted. For example, recurrent neural networks like LSTMs may be more naturally able to exploit temporal dependencies extracted by the event-graph.

\section{Acknowledgments}

This work was supported by The Horizon Europe (dAIedge, grant 101120726), the ``Excellence initiative –- research university'' programme for the AGH University of Krakow, the Polish National Science Centre projects 2024/53/N/ST6/04254 and 2024/53/N/ST6/04331 and Polish high-performance computing infrastructure PLGrid (HPC Center: ACK Cyfronet AGH -- grant no. PLG/2023/016897).

\bibliographystyle{splncs04}
\bibliography{bibtex}

\begin{thebibliography}{10}
\providecommand{\url}[1]{\texttt{#1}}
\providecommand{\urlprefix}{URL }
\providecommand{\doi}[1]{https://doi.org/#1}

\bibitem{ren2023deep}
Ren, L., Jia, Z., Laili, Y., Huang, D.: {Deep learning for time-series
  prediction in IIoT: progress, challenges, and prospects}. IEEE Transactions
  on Neural Networks and Learning Systems  (2023).
  \doi{10.1109/TNNLS.2023.3291371}

\bibitem{lim2021time}
Lim, B., Zohren, S.: Time-series forecasting with deep learning: a survey.
  Philosophical Transactions of the Royal Society A  \textbf{379}(2194),
  20200209 (2021). \doi{10.1098/rsta.2020.0209}

\bibitem{saufi2020gearbox}
Saufi, S.R., Ahmad, Z.A.B., Leong, M.S., Lim, M.H.: Gearbox fault diagnosis
  using a deep learning model with limited data sample. IEEE Transactions on
  Industrial Informatics  \textbf{16}(10),  6263--6271 (2020).
  \doi{10.1109/TII.2020.2967822}

\bibitem{dimarco2003implantable}
DiMarco, J.P.: Implantable cardioverter--defibrillators. New England Journal of
  Medicine  \textbf{349}(19),  1836--1847 (2003). \doi{10.1056/NEJMra035432}

\bibitem{Al-Ameri}
Al-Ameri, Y., Nguyen, M., Westerlund, T.: Fpga-based hardware acceleration for
  deep learning in mobile robotics. In: 2024 IEEE Nordic Circuits and Systems
  Conference (NorCAS). pp.~1--7 (2024). \doi{10.1109/NorCAS64408.2024.10752450}

\bibitem{Al-Ali}
Al-Ali, F., Gamage, T.D., Nanayakkara, H.W., Mehdipour, F., Ray, S.K.: Novel
  casestudy and benchmarking of alexnet for edge ai: From cpu and gpu to fpga.
  In: 2020 IEEE Canadian Conference on Electrical and Computer Engineering
  (CCECE). pp.~1--4 (2020). \doi{10.1109/CCECE47787.2020.9255739}

\bibitem{Guo}
Guo, K., Zeng, S., Yu, J., Wang, Y., Yang, H.: [dl] a survey of fpga-based
  neural network inference accelerators. ACM Trans. Reconfigurable Technol.
  Syst.  \textbf{12}(1) (Mar 2019). \doi{10.1145/3289185}

\bibitem{gallego2020event}
Gallego, G., Delbr{\"u}ck, T., Orchard, G., Bartolozzi, C., Taba, B., Censi,
  A., Leutenegger, S., Davison, A.J., Conradt, J., Daniilidis, K., et~al.:
  Event-based vision: A survey. IEEE Transactions on Pattern Analysis and
  Machine Intelligence  \textbf{44}(1),  154--180 (2020).
  \doi{10.1109/TPAMI.2020.3008413}

\bibitem{liu2013asynchronous}
Liu, S.C., van Schaik, A., Minch, B.A., Delbruck, T.: {Asynchronous Binaural
  Spatial Audition Sensor With 2 $\times$64$\times$4 Channel Output}. IEEE
  Transactions on Biomedical Circuits and Systems  \textbf{8}(4),  453--464
  (2013). \doi{10.1109/TBCAS.2013.2281834}

\bibitem{mostafa2024}
Mostafa, A., Hardy, E., Badets, F.: 17.8 0.4 v 988nw time-domain audio feature
  extraction for keyword spotting using injection-locked oscillators. In: 2024
  IEEE International Solid-State Circuits Conference (ISSCC). vol.~67, pp.
  328--330. IEEE (2024). \doi{10.1109/ISSCC49657.2024.10454389}

\bibitem{ortner2023online}
Ortner, T., Pes, L., Gentinetta, J., Frenkel, C., Pantazi, A.: Online
  spatio-temporal learning with target projection. In: 2023 IEEE 5th
  International Conference on Artificial Intelligence Circuits and Systems
  (AICAS). pp.~1--5. IEEE (2023). \doi{10.1109/AICAS57966.2023.10168623}

\bibitem{carpegna2024spiker}
Carpegna, A., Savino, A., Carlo, S.D.: {Spiker+: a framework for the generation
  of efficient Spiking Neural Networks FPGA accelerators for inference at the
  edge}. IEEE Transactions on Emerging Topics in Computing (01),  1--15 (Dec
  2024). \doi{10.1109/TETC.2024.3511676}

\bibitem{dalgaty2024mosaic}
Dalgaty, T., Moro, F., Demira{\u{g}}, Y., De~Pra, A., Indiveri, G., Vianello,
  E., Payvand, M.: Mosaic: in-memory computing and routing for small-world
  spike-based neuromorphic systems. Nature Communications  \textbf{15}(1), ~142
  (2024). \doi{10.1038/s41467-023-44365-x}

\bibitem{dampfhoffer_tetci_2022}
Dampfhoffer, M., Mesquida, T., Valentian, A., Anghel, L.: {Are SNNs Really More
  Energy-Efficient Than ANNs? an In-Depth Hardware-Aware Study}. IEEE
  Transactions on Emerging Topics in Computational Intelligence  \textbf{7}(3),
   731--741 (2023). \doi{10.1109/TETCI.2022.3214509}

\bibitem{dalgaty2023cnn}
Dalgaty, T., Mesquida, T., Joubert, D., Sironi, A., Soubeyrat, C., Vivet, P.,
  Posch, C.: {The CNN vs. SNN Event-camera Dichotomy and Perspectives For
  Event-Graph Neural Networks}. In: 2023 Design, Automation \& Test in Europe
  Conference \& Exhibition (DATE). pp.~1--6. IEEE (2023).
  \doi{10.23919/DATE56975.2023.10137023}

\bibitem{li2021graph}
Li, Y., Zhou, H., Yang, B., Zhang, Y., Cui, Z., Bao, H., Zhang, G.: Graph-based
  asynchronous event processing for rapid object recognition. In: Proceedings
  of the IEEE/CVF International Conference on Computer Vision. pp. 934--943
  (2021). \doi{10.1109/ICCV48922.2021.00097}

\bibitem{dalgaty2023hugnet}
Dalgaty, T., Mesquida, T., Joubert, D., Sironi, A., Vivet, P., Posch, C.:
  Hugnet: Hemi-spherical update graph neural network applied to low-latency
  event-based optical flow. In: Proceedings of the IEEE/CVF Conference on
  Computer Vision and Pattern Recognition (CVPR) Workshops. pp. 3952--3961
  (June 2023). \doi{10.1109/CVPRW59228.2023.00411}

\bibitem{mesquida2023g2n2}
Mesquida, T., Dampfhoffer, M., Dalgaty, T., Vivet, P., Sironi, A., Posch, C.:
  {G2N2: Lightweight event stream classification with GRU graph neural
  networks}. In: {https://proceedings.bmvc2023.org/}. p.~660.
  https://proceedings.bmvc2023.org/, Aberdeen, United Kingdom (Nov 2023),
  \url{https://cea.hal.science/cea-04321175}

\bibitem{jeziorek2023memory}
Jeziorek, K., Pinna, A., Kryjak, T.: Memory-efficient graph convolutional
  networks for object classification and detection with event cameras. In: 2023
  Signal Processing: Algorithms, Architectures, Arrangements, and Applications
  (SPA). pp. 160--165. IEEE (2023). \doi{10.23919/SPA59660.2023.10274464}

\bibitem{lars}
Rafeldt, L., Mesquida, T., Nakano, H., Dampfhoffer, M., Moro, F., Vivet, P.,
  Payvand, M., Dalgaty, T.: Event-based audio prediction with spectro-temporal
  event-graphs (2025)

\bibitem{jeziorek2024embedded}
Jeziorek, K., Wzorek, P., Blachut, K., Pinna, A., Kryjak, T.: {Embedded Graph
  Convolutional Networks for Real-Time Event Data Processing on SoC FPGAs}
  (2024), \url{https://arxiv.org/abs/2406.07318}

\bibitem{yang2024evgnn}
Yang, Y., Kneip, A., Frenkel, C.: Evgnn: An event-driven graph neural network
  accelerator for edge vision. arXiv preprint arXiv:2404.19489  (2024).
  \doi{10.48550/arXiv.2404.19489}

\bibitem{cramer2020}
Cramer, B., Stradmann, Y., Schemmel, J., Zenke, F.: The heidelberg spiking data
  sets for the systematic evaluation of spiking neural networks. IEEE
  Transactions on Neural Networks and Learning Systems  \textbf{33}(7),
  2744--2757 (2022). \doi{10.1109/TNNLS.2020.3044364}

\bibitem{bittar2022}
Bittar, A., Garner, P.N.: A surrogate gradient spiking baseline for speech
  command recognition. Frontiers in Neuroscience  \textbf{16} (2022).
  \doi{10.3389/fnins.2022.865897}

\bibitem{dampfhoffer2022}
Dampfhoffer, M., Mesquida, T., Valentian, A., Anghel, L.: Investigating
  current-based and gating approaches for accurate and energy-efficient
  spiking recurrent neural networks. In: Pimenidis, E., Angelov, P., Jayne, C.,
  Papaleonidas, A., Aydin, M. (eds.) Artificial Neural Networks and Machine
  Learning -- ICANN 2022. pp. 359--370. Springer Nature Switzerland, Cham
  (2022). \doi{10.1007/978-3-031-15934-3_30}

\bibitem{rossbroich2022}
Rossbroich, J., Gygax, J., Zenke, F.: Fluctuation-driven initialization for
  spiking neural network training. Neuromorphic Computing and Engineering
  \textbf{2}(4),  044016 (Dec 2022). \doi{10.1088/2634-4386/ac97bb}

\bibitem{dagostino2024}
D'Agostino, S., Moro, F., Torchet, T., Demirag, Y., Grenouillet, L., Indiveri,
  G., Vianello, E., Payvand, M.: Denram: Neuromorphic dendritic architecture
  with rram for efficient temporal processing with delays. Nature
  Communications  \textbf{15}(3446) (2024),
  \url{https://doi.org/10.1038/s41467-024-47764-w}

\bibitem{hammouamri2023}
Hammouamri, I., Khalfaoui-Hassani, I., Masquelier, T.: Learning delays in
  spiking neural networks using dilated convolutions with learnable spacings
  (2023), \url{https://arxiv.org/abs/2306.17670}

\bibitem{malettira2024}
Malettira, P.G., Negi, S., Ponghiran, W., Roy, K.: {TSkips: Efficiency Through
  Explicit Temporal Delay Connections in Spiking Neural Networks} (2024),
  \url{https://arxiv.org/abs/2411.16711}

\bibitem{sun2023}
Sun, P., Chua, Y., Devos, P., Botteldooren, D.: Learnable axonal delay in
  spiking neural networks improves spoken word recognition. Frontiers in
  Neuroscience  \textbf{17} (2023). \doi{10.3389/fnins.2023.1275944}

\bibitem{yu2022}
Yu, C., Gu, Z., Li, D., Wang, G., Wang, A., Li, E.: Stsc-snn: Spatio-temporal
  synaptic connection with temporal convolution and attention for spiking
  neural networks. Frontiers in Neuroscience  \textbf{16} (2022).
  \doi{10.3389/fnins.2022.1079357}

\bibitem{quantisenc2024}
Matinizadeh, S., Pacik-Nelson, N., Polykretis, I., Tishbi, K., Kumar, S.,
  Varshika, M.L., Mohammadhassani, A., Mishra, A., Kandasamy, N., Shackleford,
  J., Gallo, E., Das, A.: A fully-configurable open-source software-defined
  digital quantized spiking neural core architecture (2024),
  \url{https://arxiv.org/abs/2404.02248}

\bibitem{basu2022spiking}
Basu, A., Deng, L., Frenkel, C., Zhang, X.: Spiking neural network integrated
  circuits: A review of trends and future directions. In: 2022 IEEE Custom
  Integrated Circuits Conference (CICC). pp.~1--8. IEEE (2022).
  \doi{10.1109/CICC53496.2022.9772783}

\bibitem{kryjak2024event}
Kryjak, T.: Event-based vision on fpgas-a survey. In: 2024 27th Euromicro
  Conference on Digital System Design (DSD). pp. 541--550. IEEE (2024).
  \doi{10.1109/DSD64264.2024.00078}

\bibitem{xu2023event}
Xu, Y., Perera, S., Bethi, Y., Afshar, S., van Schaik, A.: Event-driven
  spectrotemporal feature extraction and classification using a silicon cochlea
  model. Frontiers in Neuroscience  \textbf{17},  1125210 (2023).
  \doi{10.3389/fnins.2023.1125210}

\bibitem{POINTNET}
Charles, R.Q., Su, H., Kaichun, M., Guibas, L.J.: Pointnet: Deep learning on
  point sets for 3d classification and segmentation. In: 2017 IEEE Conference
  on Computer Vision and Pattern Recognition (CVPR). pp. 77--85 (2017).
  \doi{10.1109/CVPR.2017.16}

\bibitem{jacob2018quantization}
Jacob, B., Kligys, S., Chen, B., Zhu, M., Tang, M., Howard, A., Adam, H.,
  Kalenichenko, D.: Quantization and training of neural networks for efficient
  integer-arithmetic-only inference. In: Proceedings of the IEEE conference on
  computer vision and pattern recognition. pp. 2704--2713 (2018).
  \doi{10.1109/CVPR.2018.00286}

\bibitem{Fey/Lenssen/2019}
Fey, M., Lenssen, J.E.: Fast graph representation learning with pytorch
  geometric. ArXiv  \textbf{abs/1903.02428} (2019),
  \url{https://api.semanticscholar.org/CorpusID:70349949}

\end{thebibliography}


%\bibitem{DVS-FPGA}
%Jeziorek, K., Wzorek, P., Blachut, K., Pinna, A., & Kryjak, T. (2024). Embedded Graph Convolutional Networks for Real-Time Event Data Processing on SoC FPGAs. arXiv preprint arXiv:2406.07318.

%\end{thebibliography}
\end{document}